# Automated Marble Plate Classification System Based On Different Neural Network Input Training Sets and PLC Implementation

Irina Topalova

Department of Automation of Discrete Production Engineering, Faculty of Mechanical Engineering
Technical University of Sofia, Bulgaria
Sofia, Bulgaria

*Abstract*— The process of sorting marble plates according to their surface texture is an important task in the automated marble plate production. Nowadays some inspection systems in marble industry that automate the classification tasks are too expensive and are compatible only with specific technological equipment in the plant. In this paper a new approach to the design of an Automated Marble Plate Classification System (AMPCS),based on different neural network input training sets is proposed, aiming at high classification accuracy using simple processing and application of only standard devices. It is based on training a classification MLP neural network with three different input training sets: extracted texture histograms, Discrete Cosine and Wavelet Transform over the histograms. The algorithm is implemented in a PLC for real-time operation. The performance of the system is assessed with each one of the input training sets. The experimental test results regarding classification accuracy and quick operation are represented and discussed.

## I. INTRODUCTION

Many modern systems for automated vision control in stone production industry are developed for inspection and classification of the surface texture of marble slabs and plates after cutting. . Many of them have high market price and are compatible only with the proprietarily company production equipment. The existing software products for texture recognition are not intended for implementation in Programmable Logic Controllers (PLC) widely used for control of technological processes. A specific task in marble plate production is the sorting of tiles with identical surface textures. The high accuracy classification in this case is difficult because the parametrical descriptions of the textures are high correlated. In addition the textures have to be recognized during movement (transportation on a conveyer belt) and with some changes in the working area illumination corresponding to the specifics of the production process. That is the reason to develop effective methods and algorithms aiming at high recognition accuracy for different kinds of similar textures when evaluating them in real production environment conditions. On the other hand optimal (considering accuracy, quick-operating and cost) software and hardware system solutions have to be sought – suitable for implementation in

PLCs, because these devices are preferable and widely used in up-to-date automated production systems. In this paper, a new approach for design of an Automated Marble Plate Classification System (AMPCS) based on different neural network input training sets is proposed, aiming at high classification accuracy using simple processing and application of only standard devices and communication protocols. It is based on training a classification Multi-Layer-Perceptron Neural Network (MLP NN) with three different input training sets: extracted texture histograms, Discrete Cosine Transform (DCT) and Discrete Wavelet Transform (DWT) over the histograms. The algorithm is implemented in a PLC for real-time operation. The system performance is assessed with each one of the input training sets. The modeling technique performance is assessed with different training and test sets. The experimental test results regarding classification accuracy and quick operation are represented and discussed.

## II. EXISTING METHODS FOR TEXTURE CLASSIFICATION

There are many contemporary methods and algorithms for texture analysis and classification. The task of classification is to associate an undefined texture with a preliminary known class. Tuceryan and Jain [1] divide the classification methods in four basic categories: statistical, geometrical, model based and filter based methods. Statistical first order data as the mean value and the dispersion are not appropriate for description of the pixel cross correlation. The second order data given by the differential statistics and the auto-correlation functions are more appropriate for texture classification [2]. The geometrical method is based on extraction of primitives, detects all the particles in the image applying operators as Sobel, Prewitt, Roberts etc. and after morphological analysis builds a detailed report on the geometrical parameters of each particle [3]. The application of statistical and filter based approaches [4] is due to the fact, that there are many software products, offering an opportunity for easy simulation and test of the results. Context analysis based on "direct local neighborhood" is also lately applied and it is a variant of a method based only on filtration. The model based analysis aims construction of a base model of the image. On the base of the constructed model a description and synthesis of the texture is feasible. The main disadvantage of the existing methods is the very high computational complexity and relatively low recognition accuracy. All of these methods are not efficient when the textures are very similar and have overlapping parametrical descriptions. When investigating recognition and classification of a preliminary known texture classes, more suitable is to apply an adaptive recognition method and a supervised learning scheme [5], since this method gives the more accurate results. In this instance the best variant is to choose neural networks (NN) because of the good NN capabilities to adapt to changes in the input vector, to set precisely the boundaries between the classes therefore offering high recognition accuracy and fast computations in the recognition phase [6]. After choosing a NN for combination between a supervised learning scheme and utilization of the histogram parameters, the more important thing is the right choice of the input NN data. This choice is influential for the right and fast convergence of the NN, for the number of parameters in the input vector and the whole NN topology. The initial choice of variables is guided by intuition. Next the number of NN input parameters has to be optimized, developing a suitable method for their reduction aiming to preserve only the informative parameters without loose of any information useful for accurate class determination [7].

## III. STRUCTURE AND FUNCTIONING MODES OF AMPCS

The AMPCS is based on training a classification MLP neural network with three different input training sets: extracted texture histograms, Discrete Cosine Transform (DCT) over the extracted histogram and Discrete Wavelet Transform (DWT) coefficients over the histogram. Many histograms of each investigated grey scale texture image, corresponding to different movement speeds of the plate on the conveyer belt and different illuminations are extracted, sampled and stored for further processing. Next DCT and DWT over the extracted histograms are also calculated and added to the stored data. The histogram values, obtained DCT and DWT coefficients are used consecutively for training a MLP neural network, changing the NN topology and the NN training parameters in order to find the best case in classification phase. The functioning structure of the designed AMPCS is shown in Figure 1. After acquisition of the marble

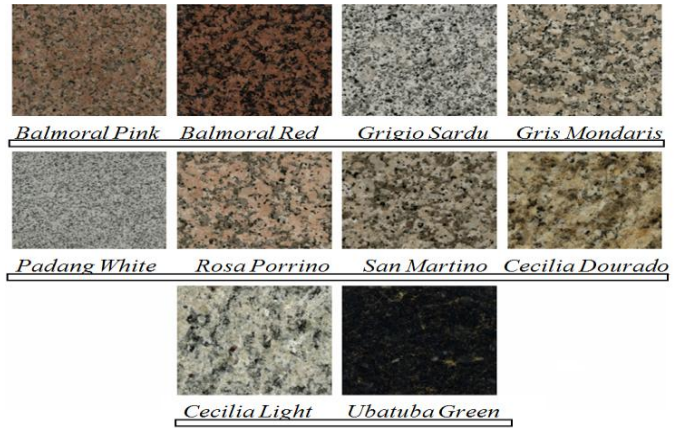

*Balmoral Pink  Balmoral Red  Grigio Sardu  Gris Mondaris*

*Padang White  Rosa Porrino  San Martino  Cecilia Dourado*

*Cecilia Light  Ubatuba Green*

Figure 2.   Exemplars of the tested marble textures.

plate images through a high resolution CCD industrial camera, they are transferred using Ethernet communication to a conventional PC [8]. The system is working in two modes – off-line or training and on-line or classification mode.

## IV. CALCULATION OF THE INPUT TRAINING SETS

### A. Histograms

The image histogram is the most important statistical characteristic. It contains information about the image contrast and brightness. Textures can also be described through their histograms. The mathematical definition of a histogram is as follows:

$$n = \sum_{i=0}^{N-1} m_i \qquad (1)$$

where for an image with a number of k grey levels, $m_i$ is each of $N = 2^k$ columns, containing the number of pixels with intensity value of i. Since the histogram values are used as NN input training vector, the values have to be fitted to the range of NN's transfer function argument - i.e. between 1 and -1. Therefore, each histogram value is divided by the maximum histogram value. To reduce NN's input layer neurons, the normalized histogram is reduced through sampling each sixth histogram value according to the method given in [9].

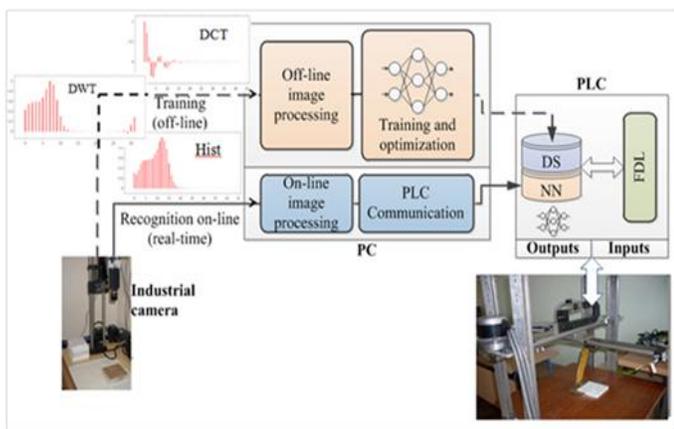

Figure 1.   Structure of the designed AMPCS.

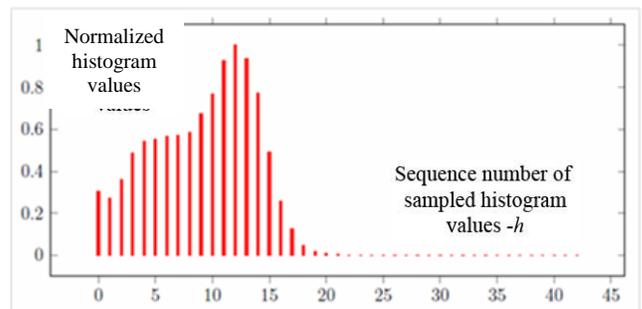

Figure 3.   Normalized and sampled histogram of marble texture *Balmoral Red*

Figure 3 shows the normalized and sampled histogram for class *Balmoral red*.

## B. Discrete Cosine Transform

Discrete Cosine Transform (DCT) represents the input signal as a sum of cosines with increasing frequencies. DCT over a signal length of N is equivalent to a Fourier transform of an even symmetrical time signal length of 2(N-1) with imaginary part equal to zero. Thus, the calculations are simplified which is a precondition for fast computations. DCT calculation over the sampled histogram h[t] is given in (2). The obtained DCT(w) is normalized with regard to the absolute maximum coefficient |DCTmax(w)|, Fig.4.

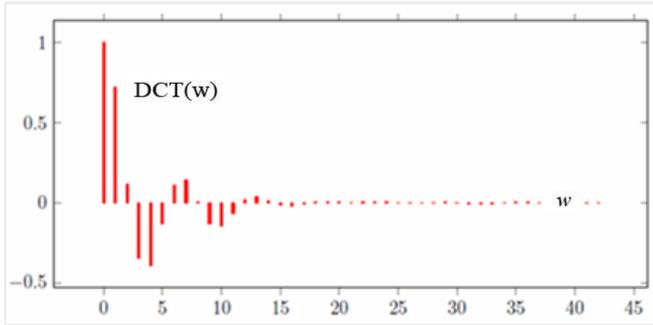

Figure 4. DCT over the histogram of the texture Balmoral Red given in Fig.3

$$DCT(w) = C_w \sum_{k=0}^{N} h[t] \cdot cos\left(\frac{(2k+1)\pi w}{2n}\right) \quad (2)$$

$w = \{0, \dots, N-1\}$ and

$$C_w = \begin{cases} \frac{1}{\sqrt{N}}, & \text{for } w = 0 \\ \sqrt{\frac{2}{N}}, & \text{for } w > 0 \end{cases}$$

## C. Discrete Wavelet Transform

Wavelet analysis consists of breaking up of signal into shifted and scaled versions of the original Wavelet $\psi(t)$. Wavelets are defined by the original wavelet function $\psi(t)$ (i.e. the mother wavelet) and scaling function $\emptyset(t)$ (also called father wavelet) in the time domain amplitudes. The most popular and used mother wavelet functions are

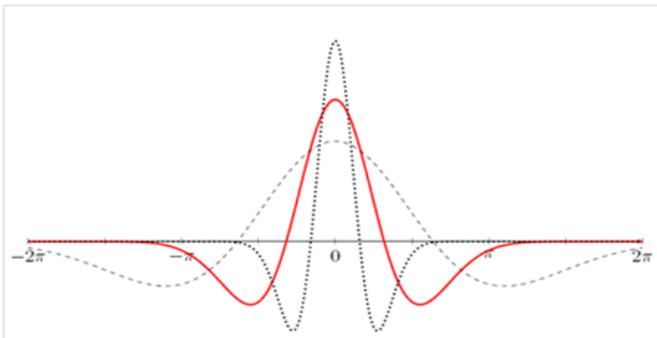

Figure 5. Three Mexican hat wavelets $\psi_{a,b}(t) = (\frac{1}{\sqrt{a}}) \psi ((t-b)/a)$ at three dilations, a = 0.5, 1.0, 2.0 and all located at b = 0.

Gaussian, Morlet, b-spline, Shannon, Mexican hat, Cohen-Daubechies-Feauveau (CDF), [10,11] etc. Here we use Mexican hat wavelet [12], Fig. 5. In this case the normalised wavelet function is often written more compactly as:

$$\psi_{a,b}(t) = \frac{1}{\sqrt{a}} \psi\left(\frac{t-b}{a}\right) \quad (3)$$

This dilation and contraction of the wavelet is governed by the dilation parameter a which, for the Mexican hat wavelet, is (helpfully) the distance between the center of the wavelet and its crossing of the time axis. The movement of the wavelet along the time axis is governed by the translation parameter b. A natural way to sample the parameters a and b is to use a logarithmic discretization of the a scale and link this, in turn to the size of steps taken between b locations. To link b to a we move in discrete steps to each location b which are proportional to the a scale [12]. This kind of discretization of the wavelet has the form:

$$\psi_{m,n}(t) = \frac{1}{\sqrt{a_0^m}} \psi\left(\frac{t-nb_0 a_0^m}{a_0^m}\right) \quad (4)$$

where the integers m and n control the wavelet dilation and translation respectively; a0 is a specified dilation step parameter set at a value greater than 1, and b0 is the location parameter which must be greater than zero[12]. Substituting a0 = 2 and b0 = 1 into equation (4) we see that the dyadic grid wavelet can be written in a very compact form as:

$$\psi_{m,n}(t) = 2^{-m/2}\psi\left(2^{-m}t - n\right) \quad (5)$$

Using the dyadic grid wavelet, the discrete wavelet transform (DWT) can be written using wavelet (detail) T m,n coefficients as:

$$T_{m,n} = \int_{-\infty}^{\infty} h(t)\,\psi_{m,n}(t)dt \quad (6)$$

The scaling function can be convolved with the signal to produce approximation coefficients as follows:

$$S_{m,n} = \int_{-\infty}^{\infty} h(t)\,\emptyset_{m,n}(t)dt \quad (7)$$

In our case the discrete input signal S0,n = h[t] is the gray scale image histogram of finite length N=256 and M=8, which is an integer power of 2: N = 2M . Thus, the range of scales we can investigate is 0 < m < M. We can represent the histogram signal h[t] using a combined series expansion using both the approximation coefficients Sm,n and the wavelet (detail) Tm,n coefficients as follows [12]:

$$h[t] = \sum_{n=-\infty}^{\infty} S_{m_o,n}\,\emptyset_{m_o,n}(t) + \sum_{m=-\infty}^{m_0} \sum_{n=-\infty}^{\infty} T_{m,n}\,\psi_{m,n}(t) \quad (8)$$

## D. Applied Wavelet multi-resolution calculation algorithm

Once we have our discrete input signal S0,n, we can compute S m,n and T m,n. This can be done for scale indices m > 0, up to a maximum scale determined by the length of the input signal. To do this, we use an iterative procedure as follows.

First we compute S1,n and T1,n from the input coefficients S0,n, i.e.

$$S_{1,n} = \frac{1}{\sqrt{2}} \sum_k c_k \, S_{o,2n+k} \qquad (9)$$

$$T_{1,n} = \frac{1}{\sqrt{2}} \sum_k b_k \, S_{o,2n+k} \qquad (10)$$

Here the scaling equation (or dilation equation) describes the scaling function $\emptyset(t)$ in terms of contracted and shifted versions of itself as follows:

$$\emptyset(t) = \sum_k C_k \, \emptyset(2t - k) \qquad (11)$$

where $\emptyset$ (2t - k) is a contracted version of $\emptyset$ (t) shifted along the time axis by an integer step k and factored by an associated scaling coefficient, ck.

Integrating both sides of the above equation, we can show that the scaling coefficient must satisfy the following constraint:

$$\sum_k C_k = 2 \qquad (12)$$

In addition, in order to create an orthogonal system we require that

$$\sum_k C_k C_{k+2k'} = \begin{cases} 2 \; if \; k' = 0 \\ 0 \; otherwise \end{cases} \qquad (13)$$

The reconfigured coefficients used for the wavelet function are written more compactly as:

$$b_k = (-1)^k C_{N_{k-1-k}} \qquad (14)$$

Next, applying the Haar filtering algorithm [10,11,12,13,14] we can find S2,n and T2,n from the approximation coefficients S 1,n and so on, up to those coefficients at scale index M, where only one approximation and one detail coefficient is computed: SM,0 and TM,0. At scale index M we have performed a full decomposition of the finite-length, discrete input signal. We are left with an array of coefficients: a single approximation coefficient value, SM,0, plus the detail coefficients, Tm,n, corresponding to discrete waves at scale a = 2m and location b = 2mn. The finite time series is of length N = 2M.

This gives the ranges of m and n for the detail coefficients as respectively 1 < m < M and 0 < n < 2 M-m - 1[12]. At the smallest wavelet scale, index m = 1, 2M=21 = N/2 coefficients are computed, at the next scale 2M=22 = N/4 are computed and so on, at larger and larger scales, until the largest scale (m = M) where only one (= 2M=2M) coefficient is computed. The total number of detail coefficients for a discrete time series of length N = 2M is then, 1 + 2 + 4 + 8 + … + 2 M-1 = N - 1. In addition to the detail coefficients, the single approximation coefficient SM,0 remains.

As the histogram S0,n = h[t] input signal vector contains N=256 components, the decomposition of the signal in our case corresponds to a number of 32 approximation coefficients Sm,n when stopped at subsequent level m=3.

The obtained S3,n approximation coefficients for the texture Balmoral Red are illustrated in Fig.6.

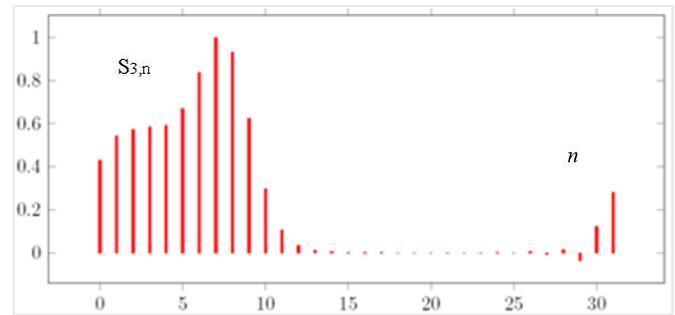

*Figure 6.* Three-stage (m=3) approximation DWT coefficients $S_{3,n}$ for texture *Balmoral Red*

## V. EXPERIMENTS AND RESULTS

The experiments were carried out with the ten classes of marble textures shown in Figure 2. The image acquisition was accomplished using a CCD camera Basler scA1000-20gm with exposure time of 120µs to 29.5 ms and resolution 1034 x 779 pixels. The value of 9Pix Motion Blur (MB), corresponding to an image resolution of 300 dpi or 118 Pix/cm, 25 m/min linear velocity of the conveyer belt, and 1/500 sec camera exposure time was applied to the images. Also 10% brightness variations were added to some of the texture samples. The MLP NN structure was trained in off-line mode of AMPCS, consecutively with sampled histogram values (each 8th value was sampled) of the ten textures, with their DCTs, sampled on the same way, and with DWTs approximating coefficients S3,n. Thus, all MLP NN input vectors are composed of 32 components. The right choice of MLP NN topology (number of layers, number of neurons in the layers) and NN parameters (activation function, MSE - mean square error, input data) is decisive for optimal training the NN [15]. Initially we use three - layered 32-50-10 MLP NN topology, where 10 is the number of output layer neurons, corresponding to the number of tested texture classes. With the training of MLP NN we want to obtain "softer" transitions or larger regions, where the output stays near to "1" or "-1" (using tangent hyperbolic as activation function).

With 3D graphic presentation it is possible to view the input/output characteristic of two inputs and one output of the neural network in spatial representation, Fig.7. After fixing the number of input values, the number of neurons in the hidden layer and applying the method given in [16], the training is repeated with reduced MSE. These reduction aims at establishing the boundaries between the classes more precisely. After that the 3D-surfaces are observed, repeating the step and reducing MSE again until the 3D-surfaces of almost all outputs have areas, where the output have regions near to "1" or "-1" as on Figure 5/b.

At this point the train phase is completed. The number of hidden neurons in MLP NN best final topology varies between 25 and 50 with MSE between 0.01 and 0.31. It was trained after 32000 iterations on the average for all 10 classes with 300 training samples per class for each of the calculated three input sets. Finally the trained NN for the three investigated input training sets is downloaded in the Programmable Logic Controller - SIMATIC S7-317 DP [17] for on-line mode operation.

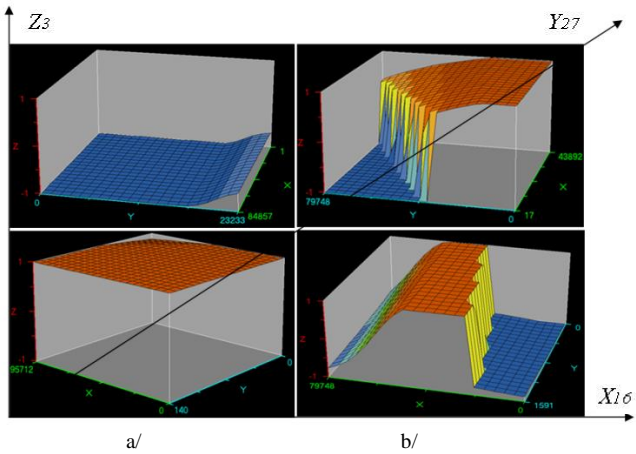

a/       b/

Figure 7. 3D surface representation of MLP NN output 3 ($Z_3$) depending on inputs № 16 ($X_{16}$) and № 27 ($Y_{27}$) of the corresponding DCTs in training phase – a/ bad training, b/ good training

These reduction aims at establishing the boundaries between the classes more precisely. After that the 3D-surfaces are observed, repeating the step and reducing MSE again until the 3D-surfaces of almost all outputs have areas, where the output have regions near to "1" or "-1" as on Fig.7/b. At this point the train phase is completed. The number of hidden neurons in MLP NN best final topology varies between 25 and 50 with MSE between 0.01 and 0.31. It was trained after 32000 iterations on the average for all 10 classes with 300 training samples per class for each of the calculated three input sets. Finally the trained NN for the three investigated input training sets is downloaded in the Programmable Logic Controller - SIMATIC S7-317 DP [18] for on-line mode operation.

## VI. Classification Performance

The algorithms for a particular neuro - application are calculated in the processor of the SIMATIC S7-317 DP and more precisely, after an unconditional call by a user program or cyclically at time-controlled intervals. MLP NN in the PLC is represented with a pair of a Function Block (reserved for NN is FB101) and an associated instance Data Block (DB101). The results of the trained NN - i.e. the obtained matrix of adapted weight coefficients are stored in the DB101. When calling FB101, NN inputs are provided with the calculated histogram, DCT or DWT values over the current accepted texture. Next a FDL-Final Decision Logic procedure for automated evaluation and finding the MLP output with maximal value (corresponding to the recognized texture class) was designed, using ladder diagram method in the PLC.

At that point we use FB1 and instance data block DB1. FB1 receives as inputs the output values of MLP NN (Fig.9). FB1 contains the developed algorithm for finding the maximum value of NN outputs. Fig.10 shows a single compare logic element (CMP), which compares one of the NN outputs with the current memorized maximum value MD201. MLP outputs together with FDL results were applied to the physical PLC outputs and to other FBs in the PLC for direct control of the sorting mechanical devices.

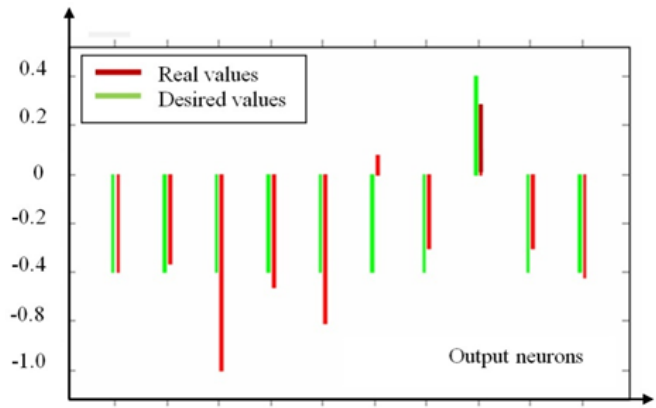

Figure 8. Output neuron values for classification of a sample of class8 – *Santa Cecilia Dourado*

Each class was tested with 40 samples of textures not included in the training set. The tests were provided with the defined three different input sets. Fig.8 shows the output neuron values for classification of a sample of class8 – Cecilia Dourado. The obtained classification accuracy, calculated as the overall number of correct classifications, divided by the number of instances in the data testset given on the average for the ten classes is presented in Table1.

The maximum value of the time needed for image acquisition, histogram/DCT/DWT calculation, giving the values to the PLC NN Function Block, classification through the trained NN and giving the result to the physical PLC outputs is calculated for different number of hidden neurons with different MSE. Fig.11 shows MLP NN outputs for correct classification of 40 test samples of class 4 - Gris Mondaris.

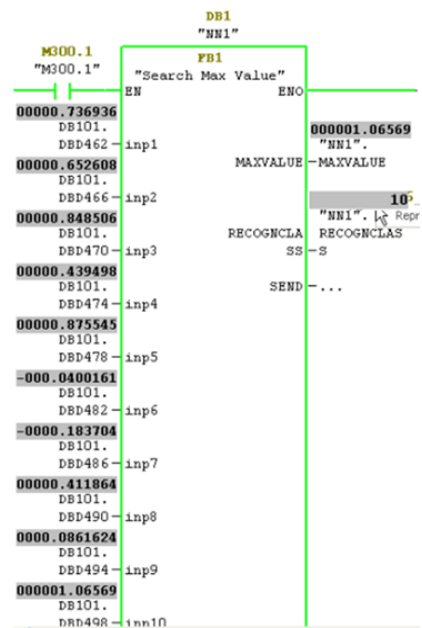

Figure 9. FDL components: Function Block of FDL

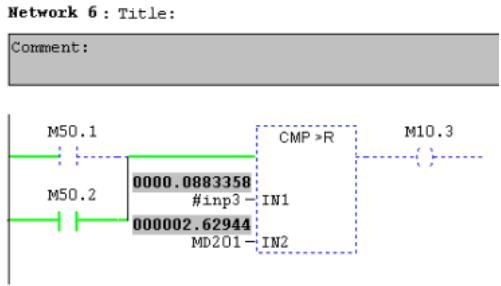

Figure 10. Compare Logic Element (CMP) in FB1

The results given in Table 1 show that the highest classification accuracy is obtained when DWT coefficients are used in AMPCS - i.e. 100% (calculated as the number of correct classified/common number of tested samples) with Mexican hat is used as mother wavelet and only 32 approximated DWT coefficients are used. When DCT coefficients are used as input set, an accuracy of 87.5 - 95% is achieved by the same length of the input vector. When a histogram with 32 sampled values is used, the accuracy is between 80% - 85%.

The best accuracy is obtained when NN has 45 hidden neurons applying the histogram method, 50 neurons – for the DCT method, and only 25 neurons for the DWT method. It is a good precondition for further minimizing the number of hidden neurons without affecting the accuracy and at the same time for reducing the real-time operation. Another good research continuation would be the testing of the accuracy and computational time applying DWT approximation coefficients with m>3 - i.e. reducing the input vector components to n=16 and even n=8. Because the faster real-time operation is achieved for the histogram method and the slowest – for the DWT method, this would compensate the relatively complicated calculations.

Finally, it is recommended to investigate the cross-correlation between texture histograms before choosing one of the proposed in this paper input sets [17]. When the cross-correlation coefficient is greater than 0.6 - 0.7, it is recommended to apply DWT method because Wavelets are building blocks that can quickly de-correlate data. In the case of slight correlated histograms (i.e. not very similar textures), it is better to apply histogram or DCT method because according to Table1 they ensure faster computation and sufficient accuracy.

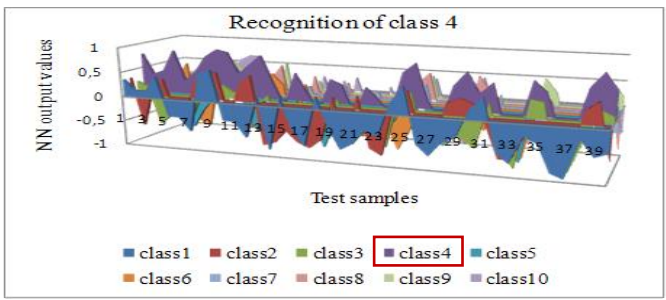

Figure 11. Classification of 40 test samples of class 4 Gris Mondaris

| Method | Number of hidden layer neurons | Mean Square Error [%] | Classification accuracy | Real-time execution [ms] |
|---|---|---|---|---|
| Histogram | 50 | 0.16 | 34/40 | 578 |
| | 45 | 0.18 | 34/40 | 553 |
| | 40 | 0.21 | 33/40 | 538 |
| | 35 | 0.24 | 33/40 | 519 |
| | 30 | 0.27 | 33/40 | 487 |
| | 25 | 0.31 | 32/40 | 461 |
| DCT | 50 | 0.01 | 38/40 | 638 |
| | 45 | 0.01 | 37/40 | 626 |
| | 40 | 0.01 | 37/40 | 611 |
| | 35 | 0.01 | 37/40 | 595 |
| | 30 | 0.01 | 35/40 | 578 |
| | 25 | 0.01 | 35/40 | 561 |
| DWT | 50 | 0.09 | 40/40 | 768 |
| | 45 | 0.10 | 40/40 | 743 |
| | 40 | 0.11 | 40/40 | 721 |
| | 35 | 0.12 | 40/40 | 698 |
| | 30 | 0.13 | 40/40 | 672 |
| | 25 | 0.16 | 40/40 | 649 |

TABLE I.    SYSTEM PERFORMANCE ACCORDING TO THE NUMBER OF HIDEN LAYER NEURONS

## II. CONCLUSION

An AMPCS using adaptive classification method was trained and tested with various marble texture images, added MB and brightness to simulate the real production conditions. It is based on training a classification MLP NN with three different input training sets: extracted texture histograms, DCT and DWT over the histograms. The developed method applying training with DWT approximation coefficients shows higher recognition accuracy compared to the other two methods and offers more resources for further reduction of the computational time while retaining the accuracy.

The proposed adaptive algorithm is implemented and proved for real-time operation on standard PLC SIMATIC S7-317 DP instead of on other embedded systems, because PLCs are widely used as control devices in automated production. Therefore, there is no need to develop additional equipment to the PLCs and this is a good opportunity for replication of the proposed AMPCS results in similar processes. The achieved short maximum system reaction time and estimated high recognition accuracy is a good precondition for proper system operation even under worse production conditions.